\pdfoutput=1

\documentclass[11pt]{article}

\usepackage[final]{coling}

\usepackage{times}
\usepackage{latexsym}
\usepackage{subfigure}
\usepackage{multirow}
\usepackage{wrapfig}
\usepackage{setspace}

\usepackage[T1]{fontenc}

\usepackage[utf8]{inputenc}

\usepackage{microtype}

\usepackage{inconsolata}

\usepackage{graphicx}

\title{Do Large Language Models Mirror Cognitive Language Processing?}

\author{Yuqi Ren\textsuperscript{\rm{1}}, 
Renren Jin\textsuperscript{\rm{1}}, 
Tongxuan Zhang\textsuperscript{\rm{2}}, 
Deyi Xiong\textsuperscript{\rm{1}}\thanks{~Corresponding author.} \\
  \textsuperscript{1} College of Intelligence and Computing, Tianjin University \\
  \textsuperscript{2} College of Computer and Information Engineering, 
  Tianjin Normal University \\
  \texttt{\{ryq20, rrjin, dyxiong\}@tju.edu.cn} \\
  \texttt{txzhang@tjnu.edu.cn}
  }

\begin{document}
\maketitle
\begin{abstract}
Large Language Models (LLMs) have demonstrated remarkable abilities in text comprehension and logical reasoning, indicating that the text representations learned by LLMs can facilitate their language processing capabilities. In neuroscience, brain cognitive processing signals are typically utilized to study human language processing. Therefore, it is natural to ask how well the text embeddings from LLMs align with the brain cognitive processing signals, and how training strategies affect the LLM-brain alignment? In this paper, we employ Representational Similarity Analysis (RSA) to measure the alignment between 23 mainstream LLMs and fMRI signals of the brain to evaluate how effectively LLMs simulate cognitive language processing. We empirically investigate the impact of various factors (e.g., pre-training data size, model scaling, alignment training, and prompts) on such LLM-brain alignment.
Experimental results indicate that pre-training data size and model scaling are positively correlated with LLM-brain similarity,\footnote{For notational simplicity, we refer to the similarity between LLM representations and brain cognitive language processing signals as LLM-brain similarity.} and alignment training can significantly improve LLM-brain similarity. Explicit prompts contribute to the consistency of LLMs with brain cognitive language processing, while nonsensical noisy prompts may attenuate such alignment. Additionally, the performance of a wide range of LLM evaluations (e.g., MMLU, Chatbot Arena) is highly correlated with the LLM-brain similarity.
\end{abstract}

\section{Introduction}
Large language models, e.g., chatGPT \citep{blog2023introducing}, have demonstrated linguistic capabilities that progressively approach human-level language comprehension and generation. The text representations learned by LLMs transform text into a high-dimensional semantic space \citep{mikolov2013distributed}. In neuroscience, brain cognitive processing signals, e.g., EEG, fMRI signals, have been widely acknowledged to be able to represent a snapshot of cognitive representations of language in human brains \citep{DBLP:conf/emnlp/XuMF16}. Such signals record human brain activities during the cognitive processing of language and play a key role in revealing the internal workings of human language cognition \citep{friederici2011brain}. Thus, it is natural to ask: how ‘‘similar'' are these two different modalities of language representations (LLMs representations and cognitive language processing signals in the brain)? Or do LLMs mirror human cognitive language processing? Answering these questions might facilitate us to interpret LLMs from the perspective of neuroscience. 

\begin{figure*}[ht]
\centering
\includegraphics[width=.75\textwidth]{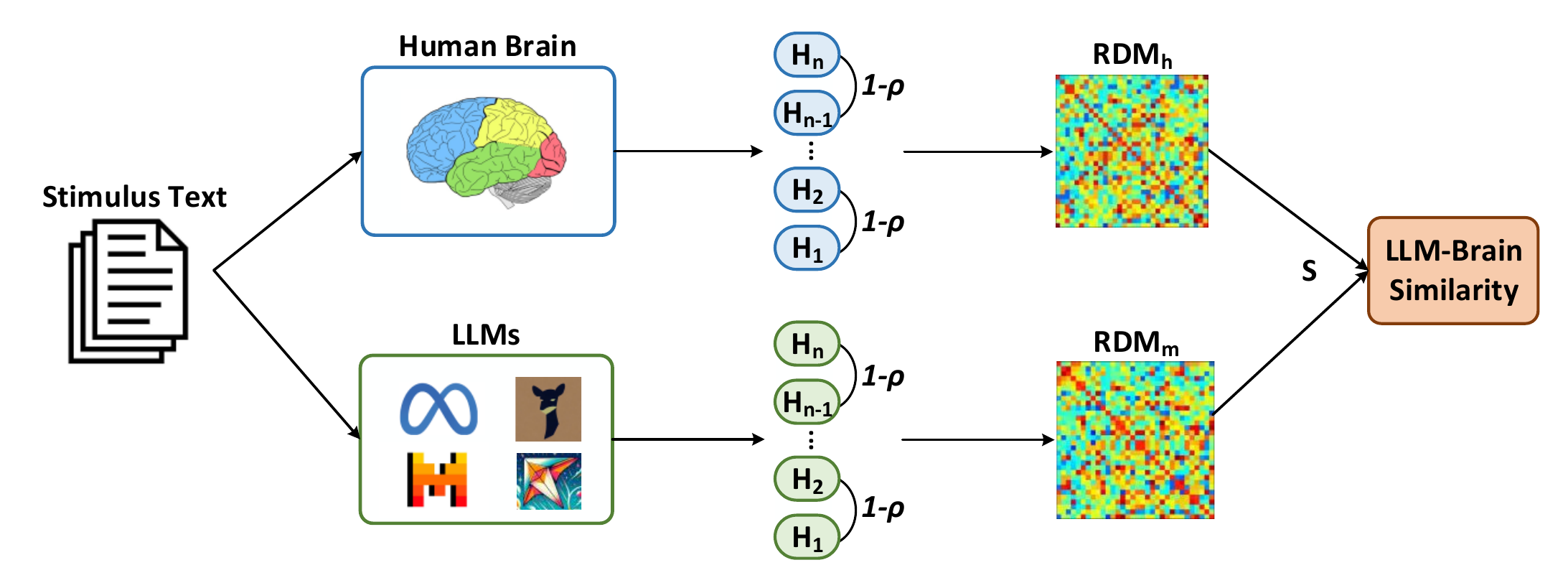}
\caption{Diagram of the proposed LLM-brain similarity estimation framework. ‘H' denotes sentence representations from different modalities. ‘$\rho$' denotes the pearson correlation coefficient. ‘S' denotes similarity measurement method. }\label{fig:model}
\end{figure*}

Previous studies have explored the assessment of consistency between representations learned by pre-trained language models and cognitive processing signals \citep{DBLP:conf/emnlp/XuMF16,DBLP:conf/conll/HollensteinTLZ19,abnar2019blackbox}. Along this research line, recent studies compare human neuroimaging data with representations learned by 43 pre-trained language models that are smaller than 1B-sized LLMs, revealing that language models with stronger predictive abilities for the next word exhibit greater alignment with the human brain \citep{schrimpf2021neural}. However, current LLMs have scaled up their model size by hundreds of times (e.g. LLaMa3-400B+ \citep{blog2024introducing}), and been strengthened by alignment training methods such as supervised fine-tuning (SFT) and reinforcement learning with human feedback (RLHF) \citep{ouyang2022training}. These factors contribute significantly to the enhanced capabilities of LLMs \citep{wei2022emergent, DBLP:journals/corr/abs-2307-04964}. Nevertheless, it is yet to be tested whether the latest LLMs with scales of 7B and above maintain consistency with human cognitive data. Additionally, the impact of pre-training data size, alignment training, prompts, and other training strategies on LLM-brain similarity has not been investigated.

This paper aims to assess the extent to which LLMs mirror cognitive language processing activity and subsequently analyze the impacts of different factors (e.g., pre-training data size, model scaling, alignment training) on such LLM-brain alignment. Specifically, we adopt functional magnetic resonance imaging (fMRI) signals as human cognitive language processing signals, and RSA \citep{kriegeskorte2008representational}, widely used in neuroscience \citep{connolly2012representation, diedrichsen2017representational}, to compute similarity scores between LLM representations and human cognitive language processing signals. The framework for LLM-brain similarity computation is illustrated in Figure \ref{fig:model}.

Our experiments comprise 23 open-source LLMs, trained with various pre-training data size, scaling and training strategies (i.e., pre-training, SFT, RLHF), to investigate the impact of these factors on LLM-brain similarity. We further analyze the impact of prompt strategy on the human intention understanding abilities of LLMs and the consistency between LLM sentiment and human sentiment. Additionally, we investigate the relationship between the capability of LLM evaluations and LLM-brain similarity.

Our main contributions can be summarized as follows.
\begin{itemize}
\item We employ human cognitive language processing signals to assess the resemblance between LLMs and human cognition. By analyzing the similarity of 23 LLMs to human cognitive language processing, we investigate the impact of various factors, such as pre-training data size, model scaling, alignment training and prompt strategy, on the LLM-brain similarity.
\item We find that the explicit-prompts result in a higher LLM-brain similarity compared to no-prompts, indicating that explicit prompts can enhance the alignment of LLMs with human intentions.
\item Our findings indicate that LLMs exhibit high similarity to human cognitive language processing in positive sentiment, suggesting that LLMs may encode more positive sentimental text during training.
\item We observe a high degree of consistency between the performance of a wide range of LLM evaluations (e.g., MMLU, Chatbot Arena) and LLM-brain similarity, indicating that the LLM-brain similarity holds substantial potential to evaluate the capabilities of LLMs.
\end{itemize}

\section{Related Work}
\paragraph{Neuroscience-Inspired NLP}
Human brain contains intricate neural networks and hierarchical structures, that encompass a broader cognitive complexity and generalization, providing a more comprehensive and profound perspective to understand the performance of LLMs in language cognition \citep{deneve2017brain, thiebaut2022emergent, ren2021cogalign}. Thus, aligning brain cognitive processing signals with text embeddings to evaluate LLMs has remained a prominent area of research. Numerous studies explore how embedding representations of words and sentences extracted from NLP models correspond to fMRI or MEG recordings \citep{mitchell2008predicting,pereira2018toward}. \citet{DBLP:conf/conll/HollensteinTLZ19} propose a framework for cognitive word embedding evaluation called CogniVal. This framework assesses the extent to which language model representations reflect semantic information in the human brain across three modalities: eye-tracking, EEG, and fMRI. \citet{DBLP:conf/emnlp/XuMF16} introduce a lightweight tool named BrainBench, designed to evaluate the word semantics in distributional models, using fMRI images corresponding to 60 concrete nouns. \citet{DBLP:conf/nips/TonevaW19} interpret the differences among 4 language models concerning layer depth, context length, and attention types using fMRI data. By leveraging insights from attention experiments, the performance of syntax-related tasks is enhanced.

\paragraph{NLP-Inspired Neuroscience}
Leveraging the superior semantic expressiveness of text embeddings from pre-trained models, a few studies utilize the computational mapping between brain activity and text embeddings to decode human brains \citep{mitchell2008predicting, pereira2018toward, ren2022bridging}. \citet{sun2019towards} have explored the effectiveness of different types of distributed representation models in sentence-level brain decoding based on fMRI, and found that supervised structured models exhibited remarkable performance in brain decoding. Additionally, \citet{wehbe2014aligning} drew an analogy between the recurrent neural network language model and the working mechanisms of the brain during discourse comprehension. They utilize the predictive performance of magnetoencephalography on hidden states at different moments to reveal what information is encoded by brain regions at specific times.

\section{Methodology}
\label{section:methodology}
To measure the extent to which LLMs align with human cognitive language processing, we employ the RSA \citep{kriegeskorte2008representational} method to assess the similarity in representations between human brains and LLMs when presented with the same stimulus text. Furthermore, we investigate how LLM-brain similarity varies with different prompts and analyze resemblances between LLMs and human cognitive language processing regarding sentimental polarity using crafted prompts.

\subsection{LLM-Brain Similarity Estimation}
RSA is a widely-used technique for comparing cross-modal representation spaces and has found extensive applications in neuroscience \citep{connolly2012representation, diedrichsen2017representational}. In NLP, this method is utilized to calculate the consistency between two different representation spaces encoding information \citep{abdou2019higher, chrupala2019correlating}. Moreover, it can be employed to quantify the relationship between human brains and neural network models \citep{DBLP:journals/neuroimage/KingGSKB19}. 

We hence use RSA to measure the LLM-brain similarity. Firstly, we compute the representation dissimilarity matrix (RDM) separately for brain cognitive processing signals and LLMs embeddings. For each pair of stimulus sentences, we calculate the 1 - Pearson correlation coefficient (${\rho}$) of their representations (either LLM representations or brain cognitive processing signals) as dissimilarity, following the previous works \citep{dwivedi2019representation, luo2022cogtaskonomy}. Consequently, both the human brain (RDM$_h$) and the LLM (RDM$_m$) yield a diagonal symmetric dissimilarity matrix of size ${n*n}$, where ${n}$ is the number of stimulus sentences. We formulate it as follows:
\begin{equation}\label{equ1}
{\rm RDM} = \left[\begin{array}{ccc}
     0 & \cdots & 1-\rho(H_1,H_n) \\
     \vdots & \ddots & \vdots\\
     1-\rho(H_n,H_1) & \cdots & 0 \\
\end{array}\right]
\end{equation}

We then estimate the similarity between RDM$_m$ and RDM$_h$ using the Pearson correlation coefficient, Euclidean distance, Cosine distance and Spearman correlation coefficient. In addition to Euclidean distance, a high correlation score implies that representations learned by the LLM are highly aligned with brain cognitive processing signals.
\begin{equation}\label{equ2}
{\rm Sim} = S({\rm RDM}_h, {\rm RDM}_m)
\end{equation}

\subsection{Prompt Strategy}
Instruction tuning significantly enhances the in-context learning ability of LLMs by understanding task requirements well through task definitions and corresponding examples \citep{ouyang2022training, peng2023instruction}. We aim to verify the effect of prompts on human intentions by examining whether appended prompts increase LLM-brain similarity.

To assess the sensitivity of the LLM-brain similarity to appended prompts, we compare the effects of three conditions: no prompts (vanilla input), explicit prompts, and noisy prompts. No-prompt-appending involves directly feeding text into LLMs, allowing the LLM to automatically continue the text without any prompts. To evaluate the effect of prompt on the same task, we prepend the phrase ‘‘Please complete the following text:'' to the input text as an explicit prompt. \begin{table}[ht]
\centering
\setlength{\tabcolsep}{20pt}
\begin{tabular}{cc}
\hline
\textbf{Positive}   & \textbf{Negative} \\
\hline
great   &    war      \\
charming    &    sin      \\
successful   &    stupid      \\
pleasure      &     prison     \\
laugh   &     pain     \\
elegance    &     liar     \\
kindness      &     angry     \\
smiling   &     damage     \\
accomplished   &    sad      \\
impress &    poor    \\
\hline
\end{tabular}
\caption{10 positive concepts vs. 10 negative concepts, serving as proxies to positive and negative sentiment. }\label{Tab:concepts}
\end{table}Furthermore, we introduce a noisy prompt consisting of 5 randomly sampled English words, e.g., ‘‘Harmony illuminate umbrella freedom like.'' to analyze the impact of nonsensical prompts under the same length.

\subsection{Sentimental Polarities}
LLMs have demonstrated significant achievements in value alignment \citep{han2023context, ganguli2023capacity}, with sentimental expressions emerging as a prominent feature in their outputs. Therefore, it is important to analyze the consistency of LLMs and human sentimental tendencies to assess the alignment ability of LLMs. Current studies typically utilize psychological scales to analyze the similarity of sentimental expression between LLMs and humans \citep{huang2023emotionally, wang2023emotional}. In this paper, we focus on the intrinsic representations of sentimental expressions in LLMs. We delve into the similarity between the embeddings of sentimental concepts learned by LLMs and brain signals under the same concepts to investigate the consistency between LLMs and humans across different sentimental polarities. It's worth noting that brain signals, which record physiological responses accompanied by emotions, are commonly used in psychology to analyze human sentiment \citep{davidson2003affective}.

Specifically, we determine the sentimental polarities (i.e., positive vs. negative) of 180 concepts in fMRI stimulus texts by the sentiment analysis tool VADER (Valence Aware Dictionary and sentiment Reasoner) included in NLTK \citep{DBLP:books/daglib/0022921}. After inputting a concept into the VADER, it will return a sentimental polarity score ranging from -1 (negative) to +1 (positive), with 0 indicating neutrality. Subsequently, we select the top 10 concepts with the highest positive sentiment and the top 10 concepts with the highest negative sentiment to serve as proxies for positive and negative sentiment, as shown in Table \ref{Tab:concepts}. To obtain LLM representations for the selected concepts, we generate prompts using the template ‘‘The sentiment of the word <concept> is'', focusing on extracting embeddings solely for the ‘‘<concept>'' token. The similarity score for positive sentiment and negative sentiment are computed by averaging the LLM-brain similarity scores across all concepts within the each polarity.

\section{Experiments}
\label{section:experiments}
We conducted experiments on 23 mainstream LLMs to investigate the alignment between LLM representations and brain signals, and the impact of various factors on this alignment. Additionally, we studied the consistency between LLMs and human cognitive language processing regarding sentimental polarity.

\subsection{Cognitive Language Processing Data}
fMRI signals map brain activity by detecting changes associated with blood flow. The fMRI data utilized in our experiments was from Pereira \citep{pereira2018toward}.\footnote{\scriptsize The dataset is publicly available at \url{https://osf.io/crwz7/}.} These fMRI signals are recorded on a whole-body 3-Tesla Siemens Trio scanner with a 32-channel head coil. The fMRI data consist of neural activity representations in cubic millimeter-sized voxels, with each fMRI recording containing a substantial number of voxels. This dataset comprises three experiments: in Experiment 1, 16 subjects read 180 concepts; in Experiment 2, 9 subjects read 384 sentences; and in Experiment 3, 6 subjects read 243 sentences. However, only 5 subjects participated in all three experiments. To ensure the fairness of our experimental results, we selected the fMRI signals from these 5 subjects. 

We flattened the 3-dimensional fMRI images into 1-dimensional vectors and randomly selected 1,000 voxels as the brain cognitive language processing signal for each subject. The stimulus texts presented to subjects consist of 180 isolated stimuli (concepts) and 627 continuous stimuli (natural language sentences). Due to individual variations among human subjects, we calculated LLM-brain similarity separately for each individual. The overall similarity scores were derived by averaging the individual scores.

\subsection{Large Language Models}
\label{subsection:LLMs}
We used 23 LLMs from 8 LLM families.
\paragraph{Amber} is a 7B LLM pre-trained on 1.3 TB of tokens, with all training codes, training parameters, and system configurations open-sourced to enhance the reproducibility and scalability of LLMs \citep{liu2023llm360}. Amber evenly divides the pre-training data into 360 data chunks, saving a checkpoint after training each chunk. Each checkpoint involves training the same amount of data as the previous one, with no differences in the number of training epochs. Additionally, several fine-tuned versions, including Amberchat and Ambersafe, have been released. Amberchat is fine-tuned using instruction data from WizardLM \citep{xu2023wizardlm}, while Ambersafe conducts direct parameter optimization (DPO) based on Amberchat to align with human preferences. In this paper, we evenly select 10 checkpoints from the publicly available 360 checkpoints to investigate the impact of pre-training data size on LLM-brain similarity.

\paragraph{LLaMA} is a series of open-source and powerful base language models, trained on trillions of tokens from publicly available datasets. We selected the 7B and 13B variants of LLaMA to investigate the impact of model scaling on LLM-brain similarity.

\paragraph{LLaMA2} is a collection of pre-trained and fine-tuned LLMs, with model size ranging from 7B to 70B \citep{touvron2023llama2}. Compared to LLaMA, the pre-training data size has increased by 40\%, and the context length has expanded from 2k to 4k. The fine-tuned version, LLaMA2-Chat, aligns with human preferences through SFT and RLHF. In this paper, we selected LLMs with sizes of 7B, 13B, and 70B from the LLaMA2 family.

\paragraph{LLaMA3} is an improved version of LLaMA2, available in both pre-trained and instruction-tuned versions \citep{blog2024introducing}. Compared to LLaMA2, LLaMA3 has made significant advancements, including the number of trained tokens has increased from 2TB to 15TB, the context window size has expanded from 4096 to 8192, and the vocabulary has enlarged from 32,000 to 128,000. During fine-tuning stage, LLaMA3 integrates SFT, rejection sampling, proximal policy optimization (PPO), and DPO. These enhancements have led to substantial improvements in inference capabilities, code generation, and instruction following. For this paper, we employed both 8B and 70B scales for the LLaMA3 family.

\paragraph{Vicuna} To investigate the effect of SFT on LLMs, we employed the Vicuna family \citep{zheng2023judging}. Vicuna v1.3 is fine-tuned from LLaMA using a training dataset comprising 125k user-shared conversations collected from ShareGPT.com. Additionally, Vicuna v1.5 is fine-tuned from LLaMA2, with the same training dataset as Vicuna v1.3. Both versions of Vicuna are available in 7B and 13B scales.
\begin{table*}[ht]
\centering
\scriptsize
\setstretch{1.3}
\begin{tabular}{clccccc}
\hline
\textbf{Scaling} & \textbf{LLM} & \textbf{Training Stage} & \textbf{Pearson} & \textbf{Euclidean} & \textbf{Cosine} & \textbf{Spearman}        \\ \hline
\multirow{12}{*}{7B}        & Amber         & Pre-training   & 0.2038 ± 0.013 & 128.26 ± 7.45 & 0.9785 ± 0.002 & 0.1440 ± 0.031          \\
                            & Amberchat     & SFT            & 0.2263 ± 0.015 & 105.41 ± 7.62 & 0.9831 ± 0.003 & 0.1477 ± 0.038          \\
                            & Ambersafe     & SFT+DPO        & 0.2302 ± 0.014 & 100.51 ± 6.93 & 0.9856 ± 0.003 & 0.1597 ± 0.032            \\  \cline{2-7}
                            & LLaMA-7B      & Pre-training   & 0.2103 ± 0.016 & 108.17 ± 6.88 & 0.9773 ± 0.001 & 0.1565 ± 0.033         \\
                            & Vicuna-7B-v1.3 & SFT            & 0.2217 ± 0.012 & 99.62 ± 7.34 & 0.9796 ± 0.005 & 0.1824 ± 0.030  \\ \cline{2-7}
                            & LLaMA2-7B      & Pre-training   & 0.2207 ± 0.016 & 105.12 ± 7.81 & 0.9800 ± 0.002 & 0.1653 ± 0.030           \\
                            & Vicuna-7B-v1.5 & SFT            & 0.2346 ± 0.014 & 98.40 ± 6.95 & 0.9850 ± 0.004 & 0.1890 ± 0.031          \\
                            & LLaMA2-7B-chat & SFT+RLHF       & 0.2410 ± 0.017 & 95.12 ± 7.44 & 0.9876 ± 0.003 & 0.1948 ± 0.034  \\ \cline{2-7}
                            & Mistral-7B     & Pre-training   & 0.2533 ± 0.013 & 100.53 ± 7.29 & 0.9767 ± 0.001 & 0.1730 ± 0.037           \\
                            & Mistral-7B-sft-alpha & SFT      & 0.2481 ± 0.016 & 93.40 ± 7.53 & 0.9808 ± 0.003 & 0.1769 ± 0.030           \\
                            & Mistral-7B-sft-beta & SFT       & 0.2573 ± 0.015 & 91.07 ± 7.66 & 0.9834 ± 0.003 & 0.1896 ± 0.032           \\
                            & Zephyr-7B       & dSFT+dDPO      & 0.2670 ± 0.014 & 89.24 ± 7.35 & 0.9876 ± 0.006 & 0.1945 ± 0.034 \\ \hline
\multirow{2}{*}{8B}         & LLaMA3-8B       & Pre-training   & 0.2540 ± 0.015 & 99.95 ± 7.62 & 0.9795 ± 0.005 & 0.1799 ± 0.035           \\
                            & LLaMA3-8B-instruct & SFT+PPO+DPO & 0.2697 ± 0.014 & 87.39 ± 7.11 & 0.9859 ± 0.002 & 0.1932 ± 0.034 \\ \hline
\multirow{5}{*}{13B}        & LLaMA-13B        & Pre-training  & 0.2221 ± 0.017 & 105.78 ± 6.86 & 0.9832 ± 0.004 & 0.1783 ± 0.033          \\
                            & Vicuna-13B-v1.3  & SFT           & 0.2427 ± 0.016 & 94.77 ± 7.49 & 0.9853 ± 0.005 & 0.1850 ± 0.037  \\ \cline{2-7}
                            & LLaMA2-13B       & Pre-training   & 0.2340 ± 0.013 & 102.50 ± 7.59 & 0.9840 ± 0.002 & 0.1799 ± 0.032          \\
                            & Vicuna-13B-v1.5  & SFT            & 0.2563 ± 0.012 & 93.68 ± 7.05 & 0.9851 ± 0.005 & 0.1776 ± 0.031           \\
                            & LLaMA2-13B-chat  & SFT+RLHF       & 0.2694 ± 0.013 & 90.89 ± 7.64 & 0.9869 ± 0.006 & 0.1831 ± 0.031  \\ \hline
\multirow{4}{*}{70B}        & LLaMA2-70B       & Pre-training   & 0.2568 ± 0.015 & 94.83 ± 7.08 & 0.9838 ± 0.002 & 0.1857 ± 0.034           \\
                            & LLaMA2-70B-chat  & SFT+RLHF       & 0.2762 ± 0.016 & 83.97 ± 6.98 & 0.9886 ± 0.003 & 0.1988 ± 0.033           \\ \cline{2-7}
                            & LLaMA3-70B       & Pre-training   & 0.2778 ± 0.015 & 92.67 ± 7.46 & 0.9877 ± 0.003 & 0.1930 ± 0.032           \\
                            & LLaMA3-70B-instruct & SFT+PPO+DPO & 0.2938 ± 0.012 & 80.94 ± 7.69 & 0.9923 ± 0.005 & 0.2098 ± 0.035  \\ \hline
\end{tabular}
\caption{The LLM-brain similarity of 23 LLMs with standard deviation. The RSA similarity calculation method includes the Pearson correlation coefficient, Euclidean distance, Cosine similarity, and Spearman correlation coefficient.}
\label{Tab:alignment}
\end{table*}

\paragraph{Mistral-7B} is a LLM with 7B parameters that outperforms the best open-source 13B model (LLaMA2) across multiple evaluation benchmarks. This model employs grouped query attention to achieve faster inference and utilizes sliding window attention to extend the processed sequence length \citep{jiang2023mistral}.

\paragraph{Mistral-7B-sft} comprises SFT versions of Mistral-7B. Mistral-7B-sft-alpha is fine-tuned on the UltraChat dataset \citep{DBLP:conf/emnlp/DingCXQHL0Z23}, which consists of 1.47M multi-turn dialogues generated by GPT-3.5-Turbo. Similarly, Mistral-7B-sft-beta is fine-tuned on a cleaned version of the UltraChat dataset, containing 200k samples.

\paragraph{Zephyr-7B} is built upon Mistral-7B-sft-beta and trained using distilled supervised fine-tuning (dSFT) and distilled direct preference optimization (dDPO) to better alignment with human intentions in interactions. This model demonstrates comparable performance to a 70B chat model aligned using human feedback \citep{DBLP:journals/corr/abs-2310-16944}.

The average embeddings of all tokens from the final layer of LLMs were utilized as sentence representations. All the LLMs used in this paper are available from \url{https://huggingface.co/models}.

\subsection{Results}
\label{section:results}
\begin{figure}[ht]
\centering
\includegraphics[width=.45\textwidth]{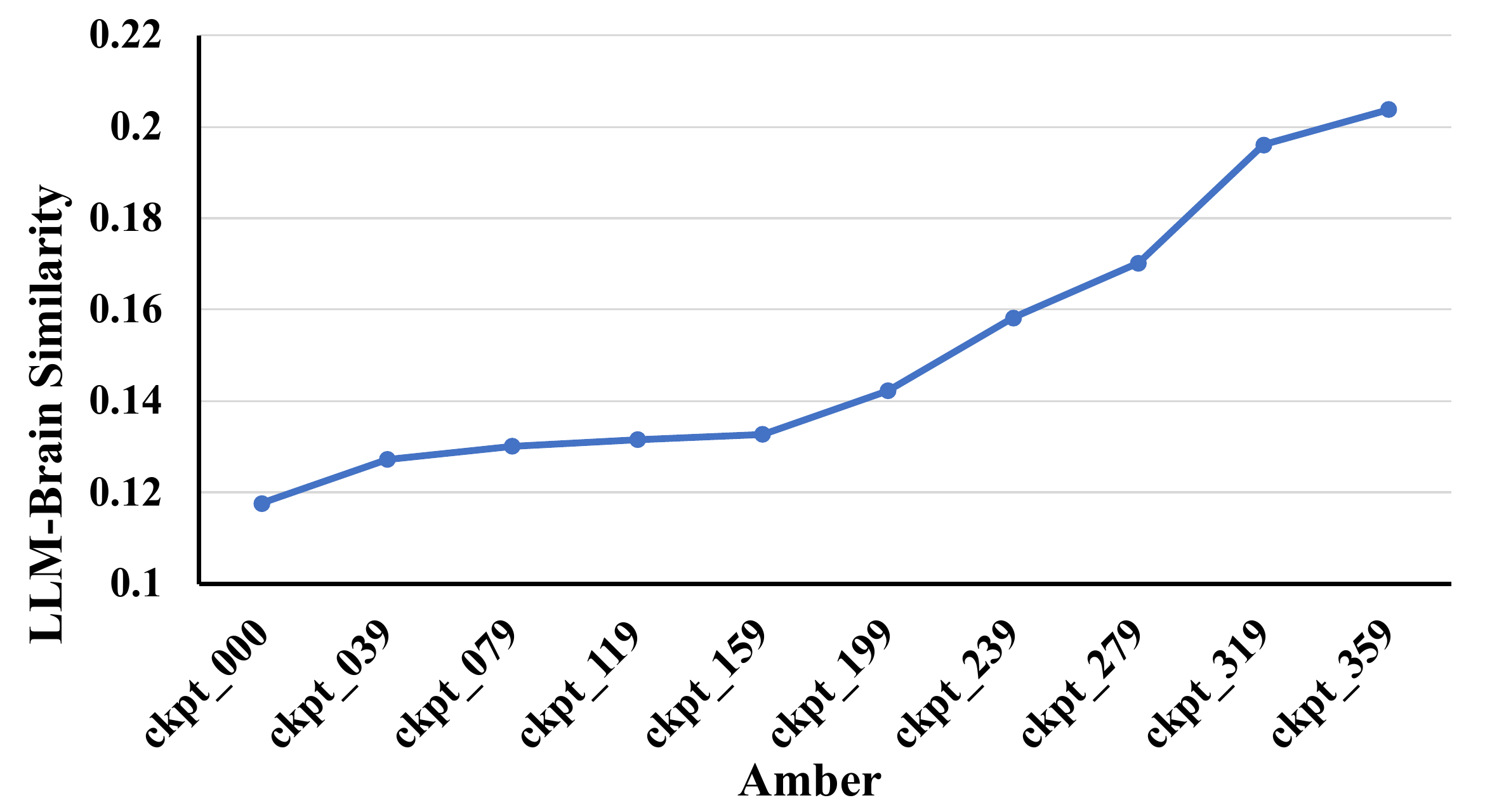}
\caption{The LLM-brain similarity of 10 different checkpoints on Amber. ‘‘ckpt'' is the abbreviation for ‘‘checkpoint''.}\label{fig:datasize}
\end{figure}
The LLM-brain similarity results for 23 LLMs are shown in Table \ref{Tab:alignment}. We observe that under different similarity measurement methods, the trend of LLM-brain similarity remains roughly consistent. This indicates that our method demonstrates robustness across different similarity evaluation methods. 

\paragraph{Pre-training data size}
The scaling law of LLMs plays a crucial role in the performance of LLMs \citep{kaplan2020scaling, hoffmann2022training}. \citet{hoffmann2022training} suggest that with a fixed scaling of LLMs, the loss gradually decreases as the amount of pre-training data increases, thereby enhancing the performance of LLMs. In this paper, we delve into how the pre-training data size affects LLM-brain similarity scores. The LLM-brain similarity scores for Amber across 10 checkpoints representing different pre-training data sizes are shown in Figure \ref{fig:datasize}, with the scores calculated by Pearson correlation coefficient. It can be observed that as the size of pre-training data increases, the LLM-brain similarity scores demonstrate gradually improvement. This indirectly suggests a negative correlation between the loss of LLMs and LLM-brain similarity scores.

\paragraph{Scaling of LLMs}
The LLM-brain similarity scores across different sizes of LLMs are illustrated in Table \ref{Tab:alignment}. We observe that the LLM-brain similarity increases with the size of LLMs. For example, when LLaMA2 is scaled from 7B to 70B, the corresponding LLM-brain similarity calculated by Pearson correlation coefficient increases from 0.2207 to 0.2568. This suggests that larger LLMs better mirror cognitive language processing of the brain than smaller LLMs. 

\paragraph{Alignment of LLMs}
Alignment training can significantly enhance the alignment capability of LLMs with human values \citep{ouyang2022training}. Therefore, we want to explore whether LLM alignment brings changes in the representation space of LLMs to narrow the gap with representations of the brain cognitive language processing. The LLM-brain similarity results for different training stages are presented in Table \ref{Tab:alignment}. From these results, we observe that:
\begin{itemize}
\item LLMs with alignment training exhibit higher similarity to brain cognitive processing signals compared to the pre-trained versions, with LLaMA3-70B-instruct achieving the highest performance. This indicates that alignment training guides LLMs to better align with human brain.
\begin{figure}[ht]
\centering
\includegraphics[width=.45\textwidth]{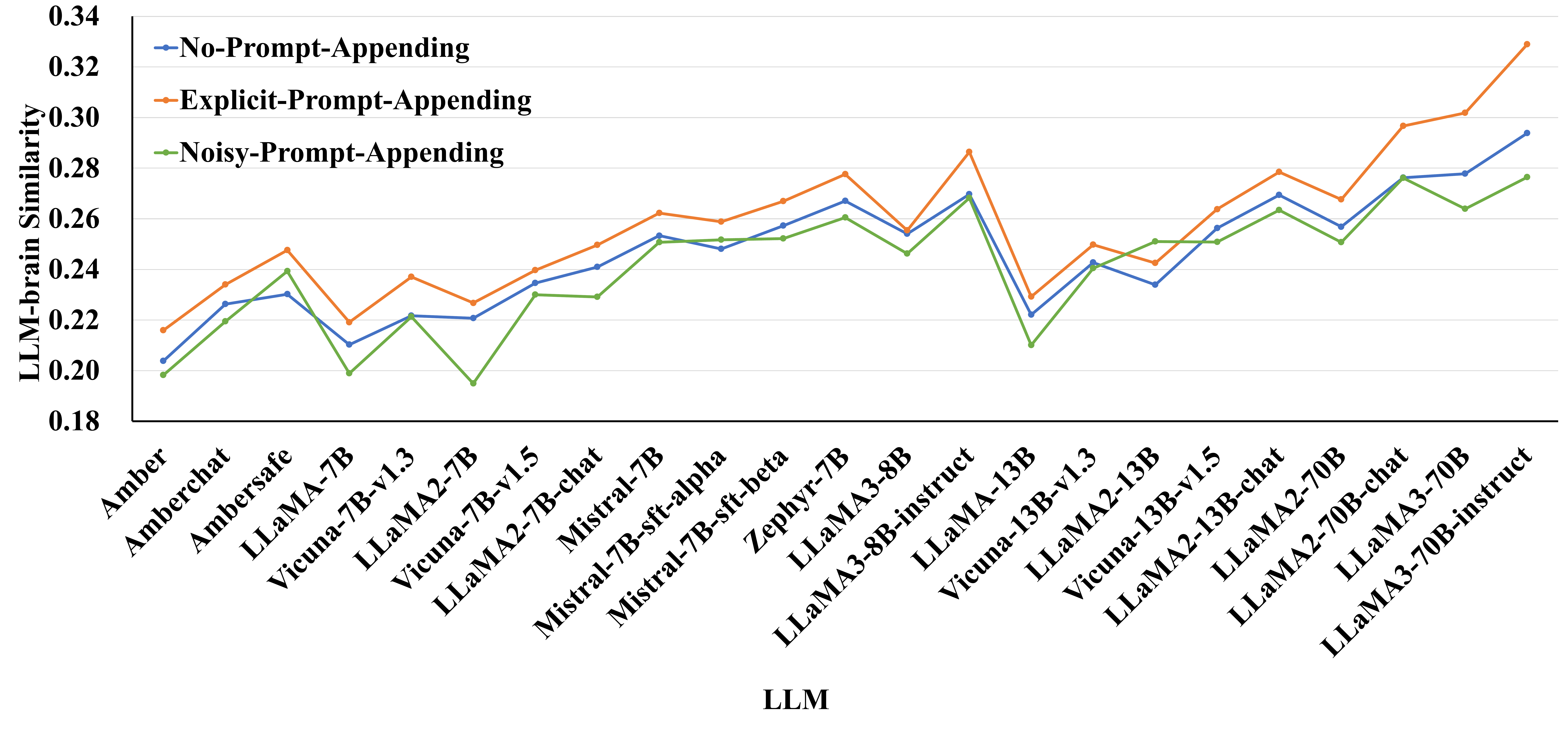}
\caption{The LLM-brain similarity calculated by Pearson correlation coefficient of different prompt appending strategies on LLMs.}\label{fig:prompt}
\end{figure}
\item Some LLMs, after alignment training, approach or surpass larger pre-trained LLMs, such as the LLaMA2-13B-chat exceeds the LLaMA2-70B by 0.0126 in Pearson correlation coefficient. This suggests that alignment training has a more significant impact on the LLM-brain similarity than model scaling.

\item Mistral-7B-sft-alpha (M-alpha) and Mistral-7B-sft-beta (M-beta) are both derived from Mistral-7B through SFT. Notably, M-alpha demonstrates lower LLM-brain similarity than Mistral-7B (0.2481 vs. 0.2573). While both models utilize SFT data from the same dataset, M-beta employs a reduced dataset by 7 fold, containing only high-quality, filtered samples. This indicates that for aligning with humans, the quality of SFT data is more important than its quantity, resonating with conclusions drawn in prior research \citep{touvron2023llama2}.
\end{itemize}

\paragraph{Sensitivity of Prompts}
The LLM-brain similarity results under the three different prompt appending strategies are illustrated in Figure \ref{fig:prompt}. It is evident that explicit-prompt-appending generally outperforms the other two types of prompt appending strategies. While the difference between no-prompt-appending and noisy-prompt-appending is relatively small, no-prompt-appending tends to outperform noisy-prompt-appending for most LLMs. This suggests that explicit prompts contribute to the consistency of LLMs with brain cognitive language processing, while nonsensical noisy prompt may attenuate such alignment.

Simultaneously, across these three types of appended prompts, LLMs consistently exhibit a positive correlation between scaling and LLM-brain similarity, indicating that the relationship between scaling and LLM-brain similarity remains insensitive to appended prompts. \begin{figure}[ht]
\centering
\includegraphics[width=0.45\textwidth]{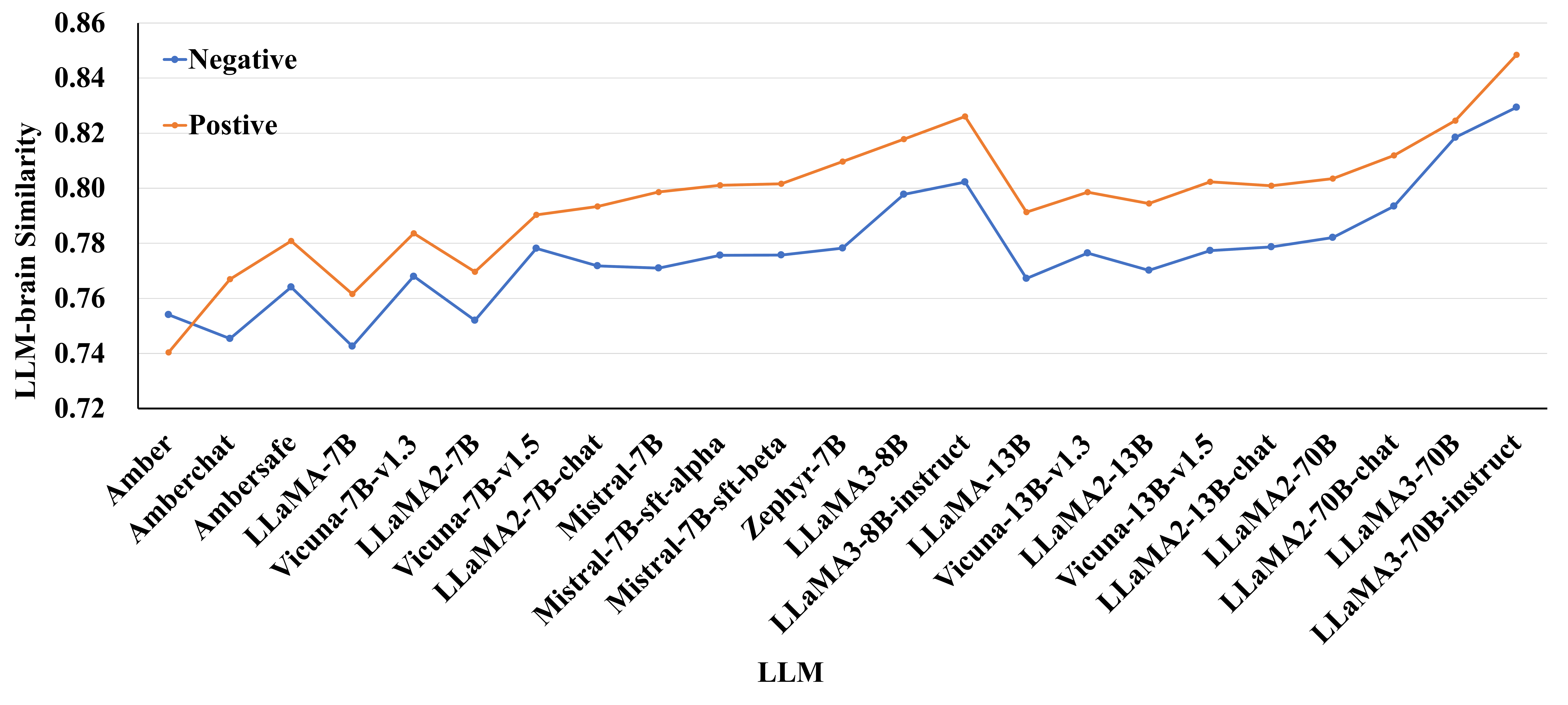}
\caption{The LLM-brain similarity calculated by Pearson correlation coefficient of LLMs across different sentimental polarities.}\label{fig:sentiment}
\end{figure}Furthermore, we explored the sensitivity of different training stages to appended prompt. We calculated the difference between the explicit-prompt-appending and no-prompt-appending, as shown in Appendix \ref{section:prompt}. It is observed that, the alignment version of LLMs demonstrates a more significant improvement in LLM-brain similarity after introducing explicit prompts compared to the pre-trained version. This might be because prompt helps LLMs align more closely with human intentions, thereby enhancing LLMs's sensitivity to prompts.


\subsection{Consistency with Human Sentiment}
The pre-training data for LLMs are predominantly from the Internet, a realm abundant with a plethora of subjective texts infused with sentimental nuances. Thus, LLMs may inherently contain sentiment-yielding capabilities, and it is worth exploring whether the sentiment of LLMs is consistent with human moral sentiment. To probe the disparities in sentimental expressions between LLMs and humans, we conducted an analysis of the LLM-brain similarity across various sentimental polarities. 

The experimental results are illustrated in Figure \ref{fig:sentiment}. We observe that the LLM-brain similarity in positive sentiment is significantly higher than that in negative sentiment. This may be due to the data governance of training data in LLMs \citep{touvron2023llama2, liu2023llm360}, such as toxicity filtering \citep{friedl2023dis,gargee2022analyzing}, which retains high-quality and positive sentiment text, resulting in a deeper encoding of positive sentiment compared to negative sentiment by the LLMs. Furthermore, alignment training enhances LLM-brain similarity in both positive and negative sentiments, indicating that LLMs encode human sentimental tendency when aligning with human sense of worth.
\begin{figure*}[ht]
\centering
\subfigure[MMLU]
{
\includegraphics[width=.45\textwidth]{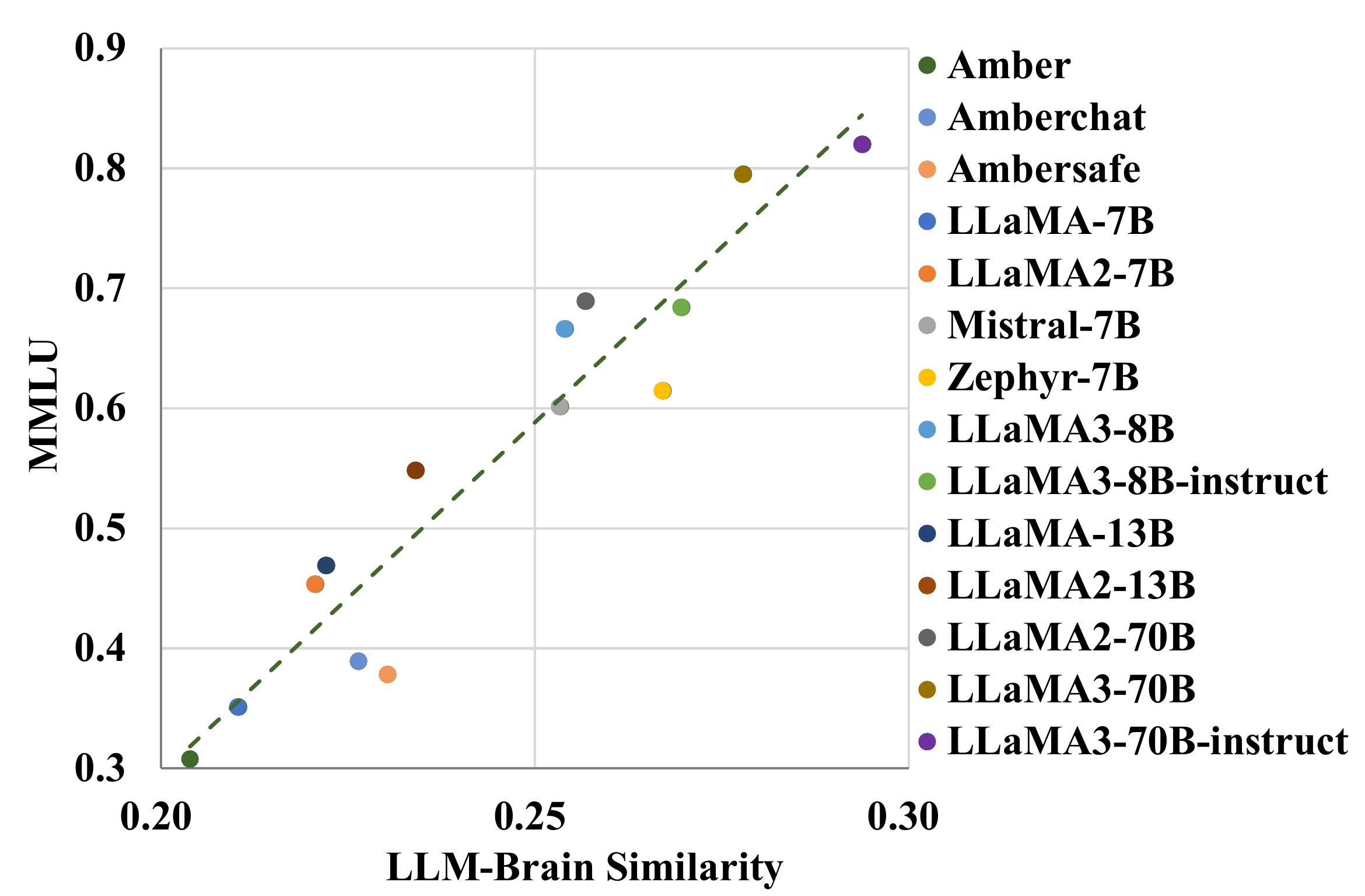}
\label{fig:MMLU}
}
\subfigure[HellaSwag]
{
\includegraphics[width=.45\textwidth]{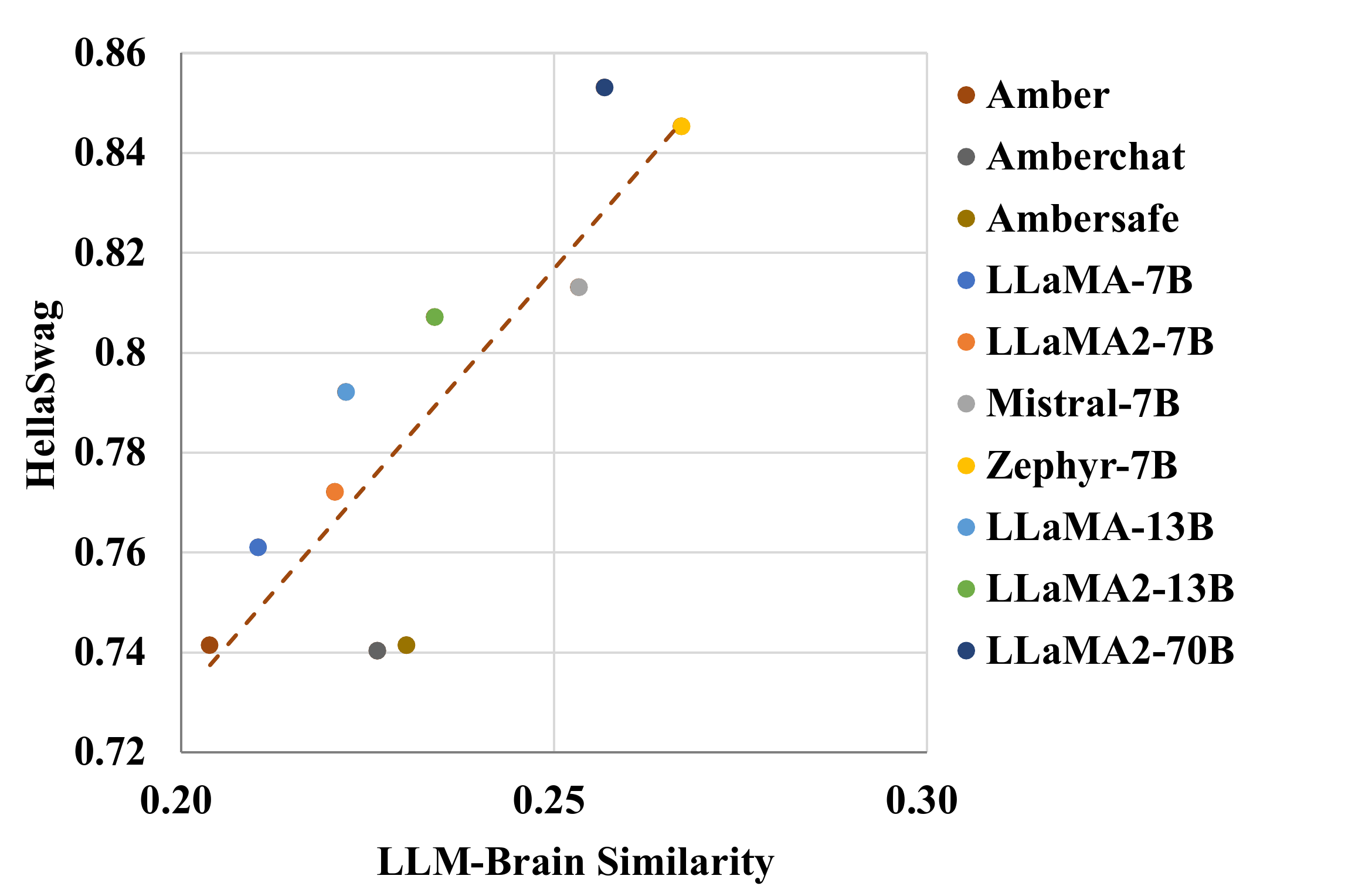}
\label{fig:hella}
}
\subfigure[Chatbot Arena ELO Rating]
{
\includegraphics[width=.45\textwidth]{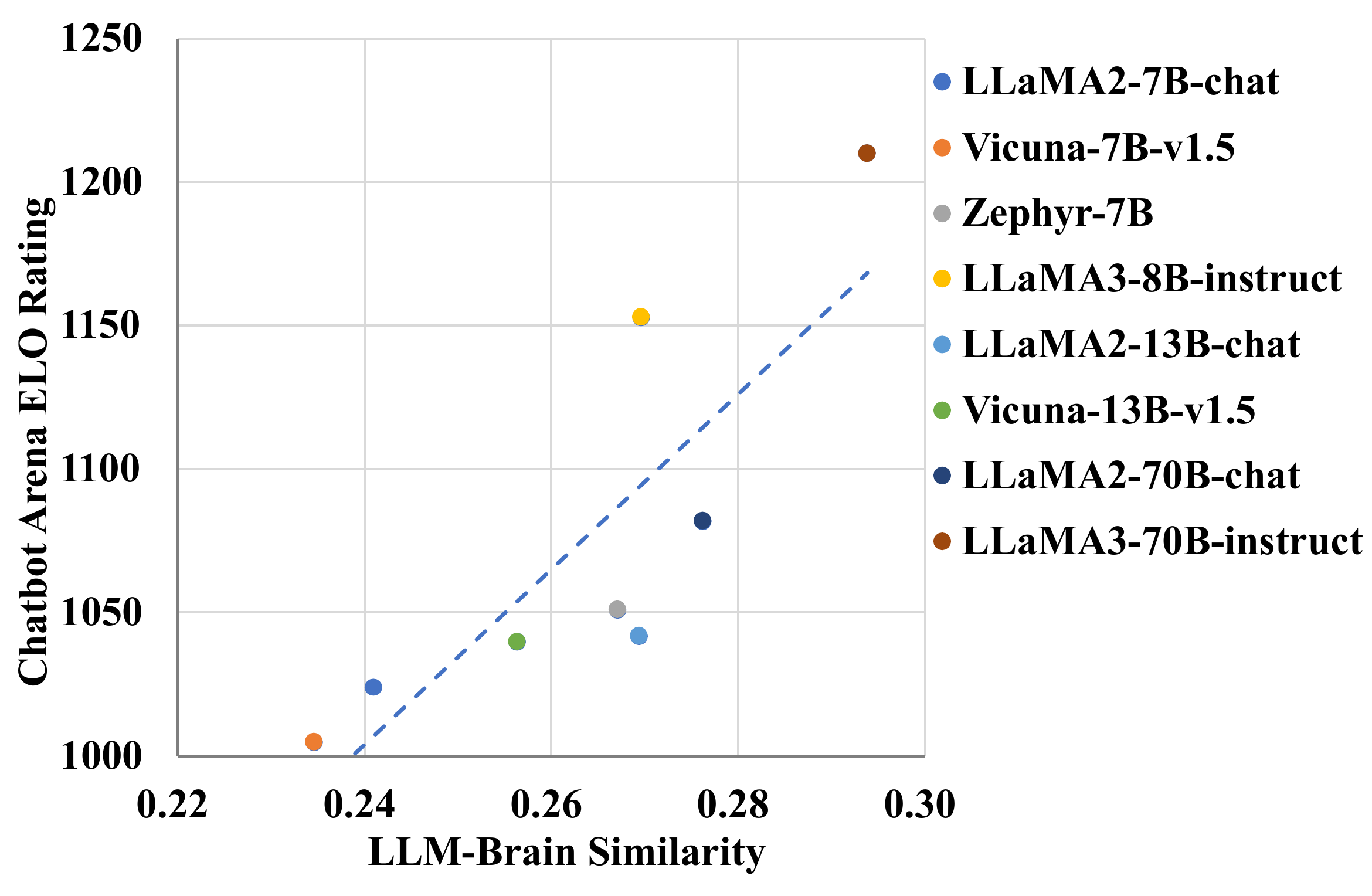}
\label{fig:chat}
}
\subfigure[AlpacaEval]
{
\includegraphics[width=.45\textwidth]{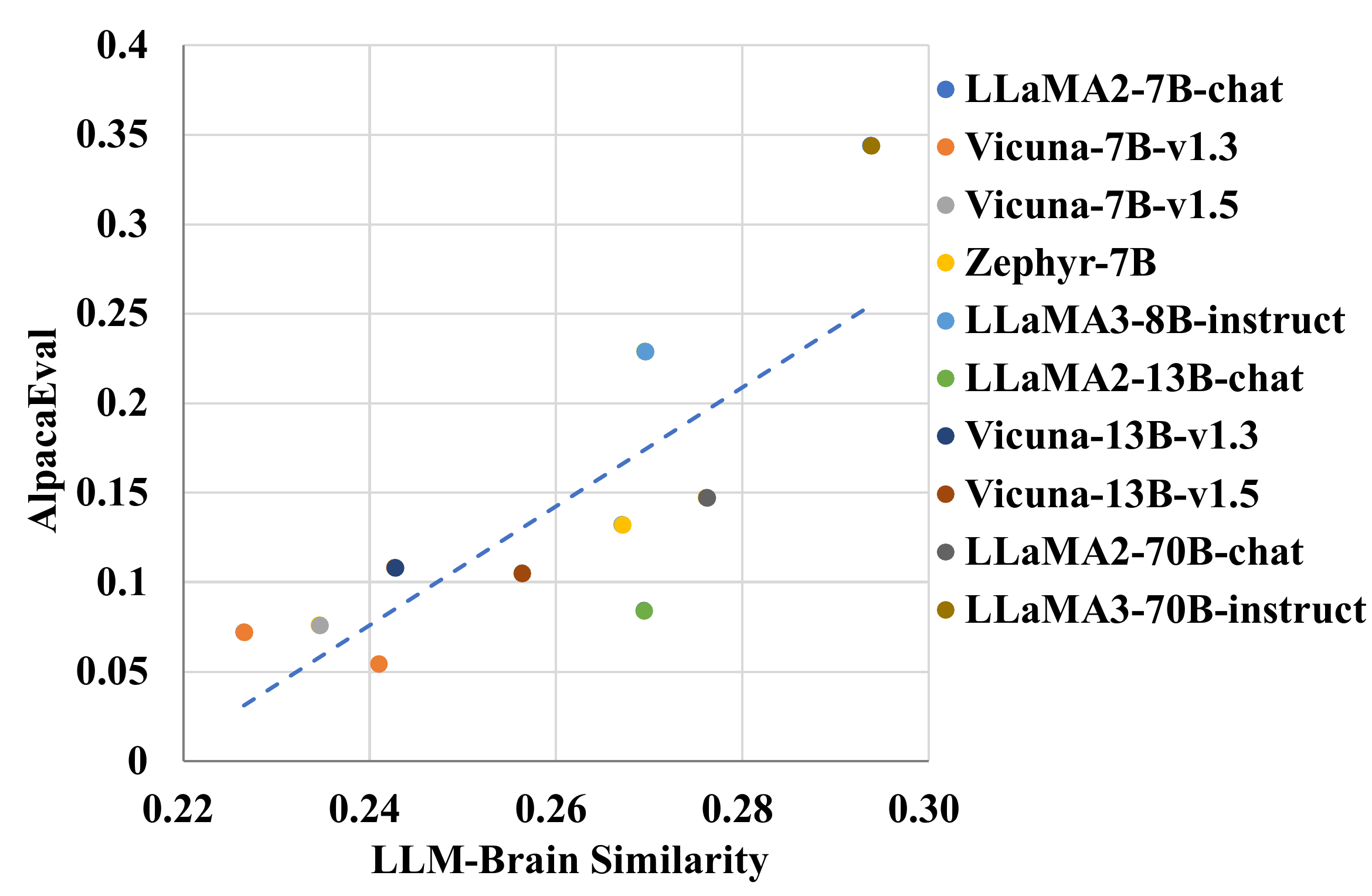}
\label{fig:alpaca}
}
\caption{Correlation between the performance of LLMs on evaluations and the LLM-brain similarity calculated by Pearson correlation coefficient.}\label{fig:evaluation}
\end{figure*}
\subsection{Correlation with LLM Evaluations}
We conducted a comparative analysis between LLM-brain similarity and performance on various LLM evaluations. To assess the knowledge and capability, we selected two benchmarks: Massive Multilingual Language Understanding (MMLU) \citep{DBLP:conf/iclr/HendrycksBBZMSS21} and HellaSwag \citep{zellers2019hellaswag}. MMLU spans 57 subjects, including STEM, humanities, and social sciences, designed to evaluate the knowledge reservoir of LLMs. HellaSwag requires LLMs to complete sentences, serving as an evaluation of commonsense reasoning abilities. For alignment evaluation, we employed Chatbot Arena Elo Rating \citep{zheng2023judging}, which uses crowdsourced methods for adversarial assessments of LLM outputs, and AlpacaEval 2.0 \citep{li2023alpacaeval}, which adopts powerful LLMs such as GPT-4 to evaluate LLMs. All performance results of these LLM evaluations are from publicly available benchmarks.

Figures \ref{fig:MMLU} and \ref{fig:hella} respectively illustrate the correlation between LLM-brain similarity and the two knowledge and capability evaluations. The results suggest a positive correlation between LLM-brain similarity and performance on these evaluations, aligning with similar findings from earlier studies \citep{DBLP:conf/conll/HollensteinTLZ19}. An intriguing finding is that LLM-brain similarity not only assesses the consistency between LLMs and human brain but also evaluates the capabilities of LLMs.

The correlation between LLM-brain similarity and alignment evaluations is illustrated in Figures \ref{fig:chat} and \ref{fig:alpaca}. The results reveal that, with the exception of LLaMA3-70B-instruct, the LLM-brain similarity and alignment capability of LLMs generally exhibit a positive correlation. LLaMA3-70B-instruct consistently achieves the highest performance in both LLM-brain similarity and alignment evaluations, significantly surpassing the linear trend line in the figures. This suggests that, although LLaMA3-70B-instruct excels in addressing artificially constructed alignment evaluations, its intrinsic representation still maintains a perceptible distance from the human brain.

\begin{table*}[ht]
\centering
\scriptsize
\setstretch{1.33}
\begin{tabular}{clccccc}
\hline
\textbf{Scaling} & \textbf{LLM} & \textbf{Training Stage} & \textbf{Pearson} & \textbf{Euclidean} & \textbf{Cosine} & \textbf{Spearman}        \\ \hline
\multirow{12}{*}{7B}        & Amber         & Pre-training   & 0.2121 ± 0.014 & 125.59 ± 6.96 & 0.9901 ± 0.001 & 0.1489 ± 0.029          \\
                            & Amberchat     & SFT            & 0.2426 ± 0.012 & 101.37 ± 7.54 & 0.9910 ± 0.002 & 0.1525 ± 0.031          \\
                            & Ambersafe     & SFT+DPO        & 0.2492 ± 0.013 & 98.54 ± 6.69 & 0.9925 ± 0.001 & 0.1576 ± 0.033            \\  \cline{2-7}
                            & LLaMA-7B      & Pre-training   & 0.2201 ± 0.014 & 102.70 ± 7.15 & 0.9902 ± 0.002 & 0.1443 ± 0.032         \\
                            & Vicuna-7B-v1.3 & SFT            & 0.2359 ± 0.015 & 96.23 ± 6.57 & 0.9910 ± 0.002 & 0.1587 ± 0.034  \\ \cline{2-7}
                            & LLaMA2-7B      & Pre-training   & 0.2278 ± 0.013 & 102.93 ± 6.89 & 0.9902 ± 0.003 & 0.1486 ± 0.028           \\
                            & Vicuna-7B-v1.5 & SFT            & 0.2395 ± 0.011 & 96.25 ± 7.15 & 0.9907 ± 0.001 & 0.1598 ± 0.030          \\
                            & LLaMA2-7B-chat & SFT+RLHF       & 0.2663 ± 0.013 & 96.95 ± 7.22 & 0.9926 ± 0.001 & 0.1586 ± 0.032  \\ \cline{2-7}
                            & Mistral-7B     & Pre-training   & 0.2326 ± 0.015 & 99.45 ± 7.64 & 0.9909 ± 0.002 & 0.1530 ± 0.031           \\
                            & Mistral-7B-sft-alpha & SFT      & 0.2564 ± 0.012 & 90.74 ± 6.38 & 0.9908 ± 0.003 & 0.1589 ± 0.030           \\
                            & Mistral-7B-sft-beta & SFT       & 0.2662 ± 0.014 & 89.65 ± 6.98 & 0.9924 ± 0.002 & 0.1620 ± 0.033           \\
                            & Zephyr-7B       & dSFT+dDPO      & 0.2802 ± 0.011 & 86.09 ± 7.14 & 0.9925 ± 0.003 & 0.1705 ± 0.031 \\ \hline
\multirow{2}{*}{8B}         & LLaMA3-8B       & Pre-training   & 0.2634 ± 0.015 & 94.52 ± 7.70 & 0.9908 ± 0.002 & 0.1559 ± 0.034           \\
                            & LLaMA3-8B-instruct & SFT+PPO+DPO & 0.2709 ± 0.012 & 89.44 ± 6.71 & 0.9924 ± 0.001 & 0.1684 ± 0.031 \\ \hline
\multirow{5}{*}{13B}        & LLaMA-13B        & Pre-training  & 0.2631 ± 0.013 & 104.98 ± 7.32 & 0.9914 ± 0.002 & 0.1653 ± 0.032          \\
                            & Vicuna-13B-v1.3  & SFT           & 0.2703 ± 0.012 & 90.60 ± 7.27 & 0.9914 ± 0.001 & 0.1659 ± 0.034  \\ \cline{2-7}
                            & LLaMA2-13B       & Pre-training   & 0.2642 ± 0.014 & 99.72 ± 7.30 & 0.9916 ± 0.001 & 0.1659 ± 0.030          \\
                            & Vicuna-13B-v1.5  & SFT            & 0.2823 ± 0.011 & 91.53 ± 6.45 & 0.9914 ± 0.001 & 0.1779 ± 0.033           \\
                            & LLaMA2-13B-chat  & SFT+RLHF       & 0.2774 ± 0.012 & 88.67 ± 6.68 & 0.9926 ± 0.002 & 0.1867 ± 0.030  \\ \hline
\multirow{4}{*}{70B}        & LLaMA2-70B       & Pre-training   & 0.2776 ± 0.014 & 90.34 ± 7.30 & 0.9923 ± 0.001 & 0.1793 ± 0.031           \\
                            & LLaMA2-70B-chat  & SFT+RLHF       & 0.3188 ± 0.012 & 82.70 ± 6.77 & 0.9929 ± 0.001 & 0.1982 ± 0.032           \\ \cline{2-7}
                            & LLaMA3-70B       & Pre-training   & 0.2906 ± 0.015 & 88.23 ± 7.22 & 0.9921 ± 0.001 & 0.1837 ± 0.034           \\
                            & LLaMA3-70B-instruct & SFT+PPO+DPO & 0.3377 ± 0.013 & 75.85 ± 7.38 & 0.9937 ± 0.001 & 0.2175 ± 0.030  \\ \hline
\end{tabular}
\caption{The LLM-brain similarity of 23 LLMs with standard deviation, when using the intermediate layer as text representations. The RSA similarity calculation method includes the Pearson correlation coefficient, Euclidean distance, Cosine similarity, and Spearman correlation coefficient.}
\label{Tab:intermediate_layer}
\end{table*}

\subsection{The Impact of LLM Layers}
The representations in different layers of LLMs capture different semantic information. Intermediate layers tend to focus on abstract concepts \cite{jawahar2019does}, while the final layer is primarily responsible for predicting the next token. We further analyze the LLM-brain similarity for 23 LLMs using text embeddings from the intermediate layers. The intermediate layer is selected by halving the total number of layers in the corresponding LLM. For example, for a 32-layer LLM, we select the 16th layer as the intermediate layer. The results are shown in Table \ref{Tab:intermediate_layer}, demonstrating that as the model scaling increases and alignment training is applied, the LLM-brain similarity in the intermediate layers improves, consistent with the trend observed in the final layer. This suggests that our findings are robust across different layers of LLMs.

\section{Conclusions}
This paper has presented a framework to estimate how well LLMs mirror human cognitive language processing. We have investigated the impact of pre-training data size, model scaling, alignment training, and prompts on the LLM-brain similarity, and explored the consistency of LLMs with humans in sentimental polarities. Experimental results reveal that expanding the size of pre-training data, scaling up models, and employing alignment training contribute to enhancing LLM-brain similarity. Moreover, explicit prompts aids LLMs in understanding human intentions, and alignment training enhances the sensitivity to prompts. Notably, LLMs exhibit a stronger resemblance to humans in positive sentiment. The strong correlation between LLM-brain similarity and various LLM evaluations suggests that the proposed LLM-brain similarity could serve as a new way to evaluate LLMs.

\section*{Limitations}
\label{section:limit}
This study, relying on text representations from open-source LLMs, is unable to assess the LLM-brain similarity of closed-source LLMs such as ChatGPT. We aim for future investigations to extend these findings to a broader spectrum of LLMs. Additionally, the fMRI stimulus texts used in this paper are exclusively in English, limiting the generalization of the observed relationships between LLMs and human cognition to other linguistic environments. Future endeavors should explore LLM-brain similarity in LLMs across diverse languages. 

\section*{Acknowledgments}
The present research was supported by the National Key Research and Development Program of China (Grant No. 2023YFE0116400) and the Natural Science Foundation of China (Grant No. 62306213). We would like to thank the anonymous reviewers for their insightful comments.
\bibliography{custom}

\clearpage
\appendix

\begin{table*}[ht]
\centering
\scriptsize
\setstretch{1.3}
\setlength{\tabcolsep}{20pt}
\begin{tabular}{clcc}
\hline
\multicolumn{1}{c}{\textbf{Scaling}} & \multicolumn{1}{l}{\textbf{LLM}} & \textbf{Training Stage} & \textbf{$\Delta$}      \\ \hline
\multirow{12}{*}{7B}    & Amber        & Pre-training     & 0.0121     \\
                        & Amberchat    & SFT              & 0.0077     \\
                        & Ambersafe    & SFT+DPO          & \textbf{0.0174}     \\ \cline{2-4}
                        & LLaMA-7B     & Pre-training     & 0.0088    \\
                         & Vicuna-7B-v1.3   & SFT         & \textbf{0.0153}      \\ \cline{2-4}
                        & LLaMA2-7B    & Pre-training     & 0.0061     \\
                                     & Vicuna-7B-v1.5                   & SFT                     & 0.0051               \\
                                     & LLaMA2-7B-chat                   & SFT+RLHF                & \textbf{0.0086}        \\ \cline{2-4}
                                     & Mistral-7B                       & Pre-training            & 0.0089                 \\
                                     & Mistral-7B-sft-alpha             & SFT                     & \textbf{0.0107}           \\
                                     & Mistral-7B-sft-beta              & SFT                     & 0.0096                   \\
                                     & Zephyr-7B                        & dSFT+dDPO               & 0.0106                   \\ \hline
\multirow{2}{*}{8B}                  & LLaMA3-8B                        & Pre-training            & 0.0014                   \\
                                     & LLaMA3-8B-chat                   & SFT+PPO+DPO             & \textbf{0.0168}           \\ \hline
\multirow{5}{*}{13B}                 & LLaMA-13B                        & Pre-training            & \textbf{0.0072}        \\
                                     & Vicuna-13B-v1.3                  & SFT                     & 0.0071       \\ \cline{2-4}
                                     & LLaMA2-13B                       & Pre-training            & 0.0086                \\
                                     & Vicuna-13B-v1.5                  & SFT                     & 0.0075        \\
                                     & LLaMA2-13B-chat                  & SFT+RLHF                & \textbf{0.0091}        \\ \hline
\multirow{4}{*}{70B}                 & LLaMA2-70B                       & Pre-training            & 0.0108                 \\
                                     & LLaMA2-70B-chat                  & SFT+RLHF                & \textbf{0.0204}      \\ \cline{2-4}
                                     & LLaMA3-70B                       & Pre-training            & 0.0241                 \\
                                     & LLaMA3-70B-chat                  & SFT+PPO+DPO             & \textbf{0.0351}        \\ \hline
\end{tabular}
\caption{The disparity in the LLM-brain similarity between the explicit-prompt-appending and no-prompt-appending strategy on different training stages. ‘$\Delta$' denotes the increment of the explicit-prompt-appending vs. the no-prompt-appending strategy.}
\label{Tab:prompt}
\end{table*}
\section{Appendix}
\subsection{The Sensitivity of Training Stages to Prompt Strategy}
\label{section:prompt}
To investigate the sensitivity of different training stages to prompt strategies, we compared the disparity of LLM-brain similarity between the explicit-prompt-appending and no-prompt-appending under the pre-trained LLMs and the alignment LLMs. The experimental results are shown in Table \ref{Tab:prompt}.

\end{document}